\documentclass[conference,9pt]{IEEEtran}
\IEEEoverridecommandlockouts

\usepackage{cite}
\usepackage{amsmath,amssymb,amsfonts}
\usepackage{algorithmic}
\usepackage{graphicx}
\usepackage{textcomp}
\usepackage{xcolor}
\usepackage{pfnote}
\usepackage{diagbox}
\usepackage{multirow}
\usepackage{booktabs}
\usepackage[hidelinks]{hyperref}
\usepackage[numbers,comma,square,sort&compress]{natbib}
\usepackage{xurl}
\usepackage{enumitem}
\setlist[itemize]{leftmargin=*}
\def\BibTeX{{\rm B\kern-.05em{\sc i\kern-.025em b}\kern-.08em
    T\kern-.1667em\lower.7ex\hbox{E}\kern-.125emX}}
\begin{document}

\title{TSCheater: Generating High-Quality Tibetan Adversarial Texts via Visual Similarity}

\author{\IEEEauthorblockN{Xi Cao\textsuperscript{1,2}, Quzong Gesang\textsuperscript{3,4}, Yuan Sun\textsuperscript{1,2*}, Nuo Qun\textsuperscript{3,4*}, Tashi Nyima\textsuperscript{3,4}}
\IEEEauthorblockA{\textsuperscript{1}\textit{Minzu University of China, Beijing, China} \\
\textsuperscript{2}\textit{National Language Resource Monitoring \& Research Center Minority Languages Branch, Beijing, China} \\
\textsuperscript{3}\textit{Tibet University, Lhasa, China} \\
\textsuperscript{4}\textit{Collaborative Innovation Center for Tibet Informatization by Ministry of Education \& Tibet Autonomous Region, Lhasa, China} \\
metaphor@outlook.com, gsqz1212@qq.com, sunyuan@muc.edu.cn, \{q\_nuo, nmzx\}@utibet.edu.cn}
\thanks{*Corresponding author.}
}

\maketitle

\begin{abstract}
Language models based on deep neural networks are vulnerable to textual adversarial attacks.
While rich-resource languages like English are receiving focused attention, Tibetan, a cross-border language, is gradually being studied due to its abundant ancient literature and critical language strategy.
Currently, there are several Tibetan adversarial text generation methods, but they do not fully consider the textual features of Tibetan script and overestimate the quality of generated adversarial texts.
To address this issue, we propose a novel Tibetan adversarial text generation method called \textit{TSCheater}, which considers the characteristic of Tibetan encoding and the feature that visually similar syllables have similar semantics.
This method can also be transferred to other abugidas, such as Devanagari script.
We utilize a self-constructed Tibetan syllable visual similarity database called \textit{TSVSDB} to generate substitution candidates and adopt a greedy algorithm-based scoring mechanism to determine substitution order.
After that, we conduct the method on eight victim language models.
Experimentally, \textit{TSCheater} outperforms existing methods in attack effectiveness, perturbation magnitude, semantic similarity, visual similarity, and human acceptance.
Finally, we construct the first Tibetan adversarial robustness evaluation benchmark called \textit{AdvTS}, which is generated by existing methods and proofread by humans.
\end{abstract}


\section{Introduction}

The vulnerability of deep neural network models to adversarial attacks was first identified in computer vision \cite{DBLP:journals/corr/SzegedyZSBEGF13,DBLP:journals/corr/GoodfellowSS14}.
In natural language processing, initially, Jia and Liang found that adding distracting sentences to input paragraphs can make reading comprehension systems answer incorrectly \cite{jia-liang-2017-adversarial}.
Since then, textual adversarial attack methods with different granularity (character-, word-, sentence-, and multi-level), in different settings (white- and black-box), and for different tasks (text classification, sentiment analysis, etc.) have been proposed \cite{10.1145/3593042}.
Even the state-of-the-art ChatGPT has been subject to many such attacks \cite{10.1162/dint_a_00243}.
Textual adversarial attack refers to an attack in which an attacker adds imperceptible perturbations to the input, thereby causing a change in the output of a language model.
The texts generated by textual adversarial attacks are called adversarial texts, which play an important role in security, evaluation, explainability, and augmentation \cite{chen-etal-2022-adversarial}.
Currently, most of the research focuses on English \cite{iyyer-etal-2018-adversarial,eger-etal-2019-text,garg-ramakrishnan-2020-bae,liu-etal-2022-character}, but due to differences in language resources and textual features, many methods cannot be transferred to other languages.

\begin{figure}[t]
	\centering
	\includegraphics[width=0.6\linewidth]{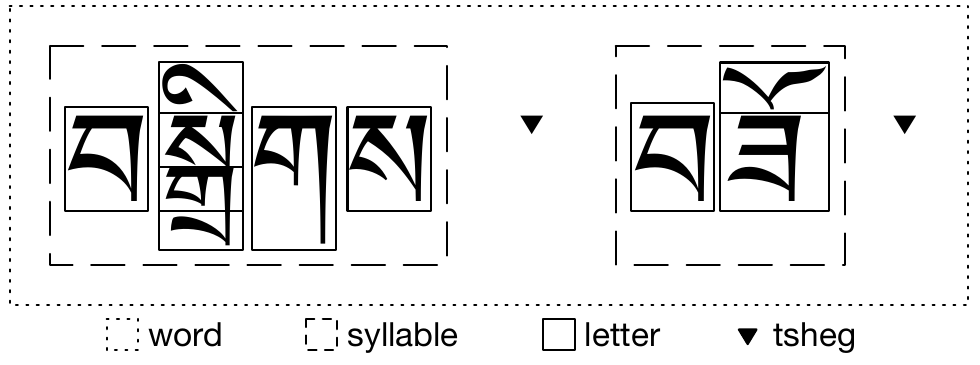}
	\caption{Structure of the Tibetan word \textit{Programming}.}
	\label{Fig.1}
\end{figure}

Tibetan is studied not only for its abundant ancient literature \cite{8978019,10.1145/3654811} but also for its critical language strategy.
It covers parts of China, India, Pakistan, Nepal, and Bhutan as a cross-border language.
Tibetan script is an abugida consisting of 30 consonant letters and 4 vowel letters, each with a corresponding Unicode.
Tibetan syllables are composed of these letters according to strict rules, and Tibetan words are composed of one or more syllables separated by Tibetan separators called tsheg.
Fig. \ref{Fig.1} clearly illustrates the structure of a Tibetan word.
Cao et al. proposed a syllable-level black-box Tibetan textual adversarial attack method named \textit{TSAttacker} to call attention to the adversarial robustness of Chinese minority language models \cite{cao-etal-2023-pay-attention}.
The method utilizes cosine distance between static syllable vectors to generate substitution syllables but does not perform well in attack granularity diversity and word embedding quality.
\textit{TSTricker} utilizes two BERT-based masked language models with tokenizers of two different granularity to generate substitution syllables or words respectively, which solves the above shortcomings and greatly improves the attack effect \cite{10.1145/3589335.3652503}.
However, these methods only evaluate the attack effectiveness and perturbation magnitude, without fully considering the textual features of Tibetan script and overestimating the quality of generated adversarial texts.

\begin{figure}[t]
	\centering
	\includegraphics[width=0.33\linewidth]{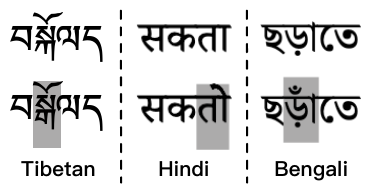}
	\caption{Visual perturbations to abugidas.}
	\label{Fig.2}
\end{figure}

In this paper, we propose a simple and sweet Tibetan adversarial text generation method called \textit{TSCheater}, which considers the characteristic of Tibetan encoding and the feature that visually similar syllables have similar semantics.
This method can also be transferred to other abugidas, such as Devanagari script.
Fig. \ref{Fig.2} depicts the visual perturbations to abugidas: Tibetan, Hindi and Bengali.
We conduct the existing methods on 8 victim language models and make a comprehensive evaluation in 5 dimensions: attack effectiveness, perturbation magnitude, semantic similarity, visual similarity, and human acceptance.
The experimental results demonstrate the superiority of \textit{TSCheater}.
Finally, we construct the first Tibetan adversarial robustness evaluation benchmark called \textit{AdvTS}, which is generated by existing methods and proofread by humans.
To facilitate future research, we make all the work in this paper publicly available at \url{https://github.com/metaphors/TSAttack}.

\section{Methodology}

\subsection{Textual Adversarial Attacks on Text Classification}

For a text classifier $F$, let $x$ ($x\in{X}$, $X$ includes all possible input texts) be the original input text and $y$ ($y\in{Y}$, $Y$ includes all possible output labels) be the corresponding output label of $x$, denoted as
\begin{equation}
	F(x)={\mathop{\arg\max}_{\dot{y}\in{Y}}{P(\dot{y}|x)}}={y}.
\end{equation}

For a successful textual adversarial attack, let $x'=x+\delta$ be the perturbed input text, where $\delta$ is the $\epsilon$-bounded, imperceptible perturbation, denoted as
\begin{equation}
	F(x')={\mathop{\arg\max}_{\dot{y}\in{Y}}{P(\dot{y}|x')}}\neq{y}.
\end{equation}

\subsection{Generate Substitution Candidates}

According to the characteristic of Tibetan encoding, the granularity of Tibetan script can be divided into letters [or Unicodes], syllables, and words, which is different from English (letters [or Unicodes], words) and Chinese (characters [or Unicodes], words).
The encoding characteristic also appears in other abugidas like Devanagari script.
For the original input text $x$, let $s$ be each syllable and $w$ be each word in it, denoted as
\begin{equation}
	x=s_{1}s_{2}\dots{s_{i}}\dots{s_{m}}=w_{1}w_{2}\dots{w_{j}}\dots{w_{n}}.
\end{equation}

According to the textual features of Tibetan script, many pairs of consonant letters look similar, and the proportion of vowel letters in syllables is small, which leads to the fact that for one syllable, there will be a number of syllables with similar appearance.
Interestingly, many abugidas also exhibit these textual features.
In addition, visually similar syllables have similar semantics and are easily produced by human typing errors, which makes them a good source of adversarial texts.
However, Tibetan syllable construction rules are complex, and it is difficult to identify all the visually similar syllables and measure their similarity with rule-based methods.
The normalized cross-correlation matching algorithm \cite{lewis1995fast} is a classical algorithm in image matching, which is often used as a representation of the matching degree or similarity degree.
CCOEFF\_NORMED\footnote{\url{https://docs.opencv.org/4.10.0/de/da9/tutorial_template_matching.html}} (Normalized Cross-Correlation Coefficient) between two images is denoted as below:
\begin{equation}
	R(x,y)= \frac{ \sum_{x',y'} (T'(x',y') \cdot I'(x+x',y+y')) }{ \sqrt{\sum_{x',y'}T'(x',y')^2 \cdot \sum_{x',y'} I'(x+x',y+y')^2} }.
\end{equation}
We adopt it to measure the visual similarity between 18,833 Tibetan syllables and construct a Tibetan syllable visual similarity database called \textit{TSVSDB}.
The closer CCOEFF\_NORMED is to 1, the more similar the two syllables are.
The construction process of \textit{TSVSDB} is as follows.
Firstly, 18,833 Tibetan syllables are converted into images using the Pillow library.
The font used is Noto Serif Tibetan\footnote{\url{https://fonts.google.com/noto/specimen/Noto+Serif+Tibetan}}, and the font size is 50.
Each image has the same width, height, black background, and white text.
The syllables are rendered from the top left corner of the images.
Then, CCOEFF\_NORMED between each syllable grayscale image and all other syllable grayscale images are calculated using the OpenCV library.
TABLE \ref{Tab.1} shows part of \textit{TSVSDB}.

Here, we segment Tibetan script into syllables by tsheg and into words using the segmentation tool TibetSegEYE\footnote{\url{https://github.com/yjspho/TibetSegEYE}}.
For each syllable $s$ in the original input text $x$, we use syllables with CCOEFF\_NORMED within $(0.8, 1)$ as the set of substitution syllables $C_s$.
For each word $w$ in $x$, firstly, we get the set of syllables with CCOEFF\_NORMED within $(0.8, 1]$ for each syllable in $w$, then perform Cartesian product operation on the sets corresponding to $w$.
After that, we filter out the elements in the Cartesian product set whose CCOEFF\_NORMED product of syllables is within $(0.8, 1)$.
The syllables in each element are concatenated into a word and these words are used as the set of substitution words $C_w$.

\begin{table}[t]
	\centering
	\caption{Tibetan Syllables Visually Similar to \includegraphics[scale=0.1]{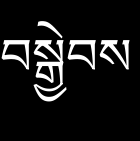}}
	\label{Tab.1}
	\begin{tabular}{m{1cm}<{\centering}m{2.5cm}<{\centering}|m{1cm}<{\centering}m{2.5cm}<{\centering}}
		\toprule
		Syllable & CCOEFF\_NORMED & Syllable & CCOEFF\_NORMED \\
		\midrule
		\includegraphics[scale=0.1]{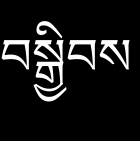} & 0.9806 & \includegraphics[scale=0.1]{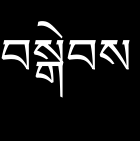} & 0.9370 \\
		\includegraphics[scale=0.1]{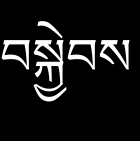} & 0.9705 & \includegraphics[scale=0.1]{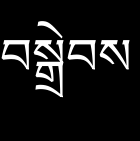} & 0.9352 \\
		\includegraphics[scale=0.1]{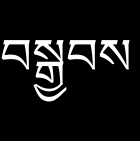} & 0.9690 & \includegraphics[scale=0.1]{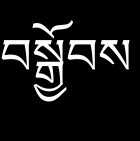} & 0.9312 \\
		\includegraphics[scale=0.1]{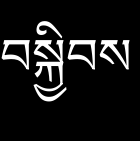} & 0.9510 & \includegraphics[scale=0.1]{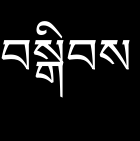} & 0.9168 \\
		\includegraphics[scale=0.1]{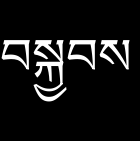} & 0.9388 & \multicolumn{2}{c}{\dots} \\
		\bottomrule
	\end{tabular}
\end{table}

\begin{figure*}[t]
	\centering
	\includegraphics[width=0.7\linewidth]{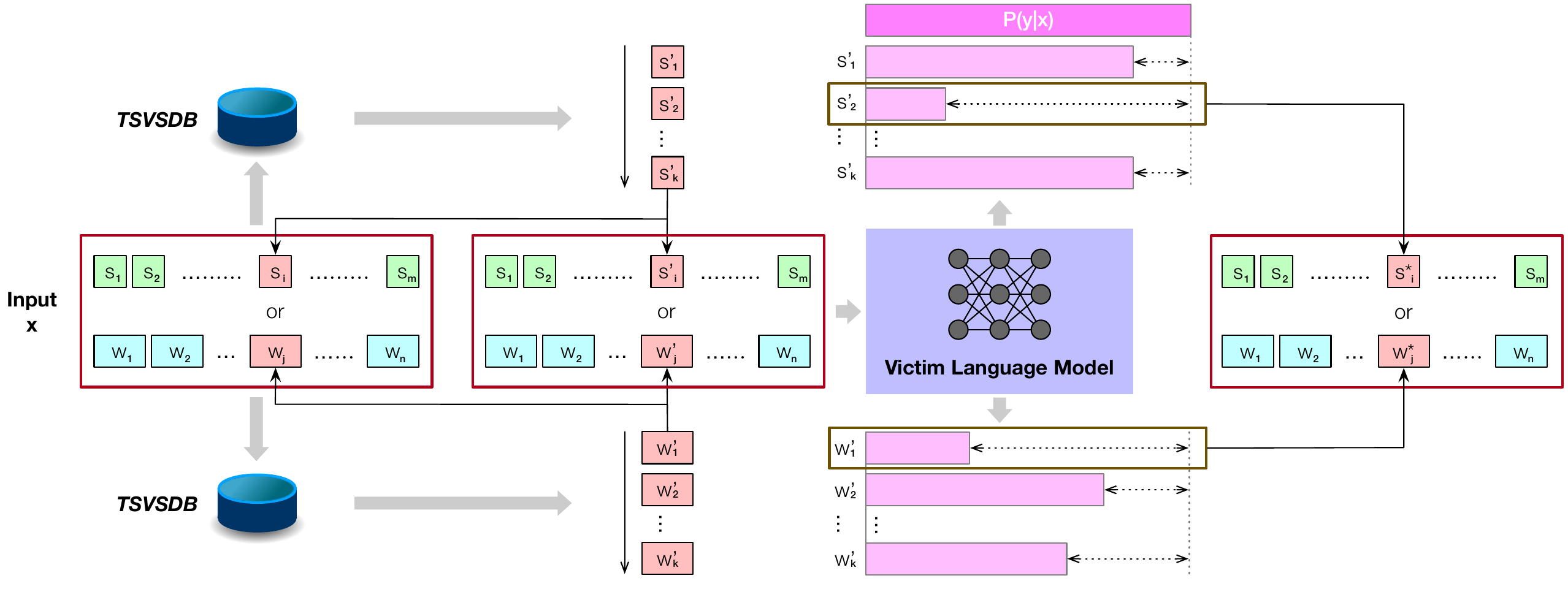}
	\caption{Process of generating substitution candidates.}
	\label{Fig.3}
\end{figure*}

Now, we iterate over each syllable $s'$ in $C_s$ or each word $w'$ in $C_w$, then
\begin{equation}
	x'=s_{1}s_{2}\dots{{s_{i}}'}\dots{s_{m}}~or~x'=w_{1}w_{2}\dots{{w_{j}}'}\dots{w_{n}},
\end{equation}
and calculate
\begin{equation}
	\Delta{P}=P(y|x)-P(y|x').
\end{equation}
After the iteration, syllable $s^*$ or word $w^*$ can always be found, and
\begin{equation}
	x^*=s_{1}s_{2}\dots{{s_{i}}^*}\dots{s_{m}}~or~x^*=w_{1}w_{2}\dots{{w_{j}}^*}\dots{w_{n}},
\end{equation}
at the moment,
\begin{align}
	\Delta{P^*}=\max\{P(y|x)-P(y|{{x'}_k})\}_{k=1}^{size(C_s~or~C_w)},
\end{align}
\begin{align}
	{s^*}~or~{w^*}=\mathop{\arg\max}_{{s'}\in{C_{s}}\atop{{w'}\in{C_{w}}}}\{P(y|x)-P(y|{x'}_{k})\}_{k=1}^{size(C_s~or~C_w)}.
\end{align}
Substituting the syllable $s$ with $s^*$ or the word $w$ with $w^*$ can cause the most significant decrease in classification probability, which has the best attack effect.
Fig. \ref{Fig.3} depicts the process of generating substitution candidates.

\begin{figure}[t]
	\centering
	\includegraphics[width=0.9\linewidth]{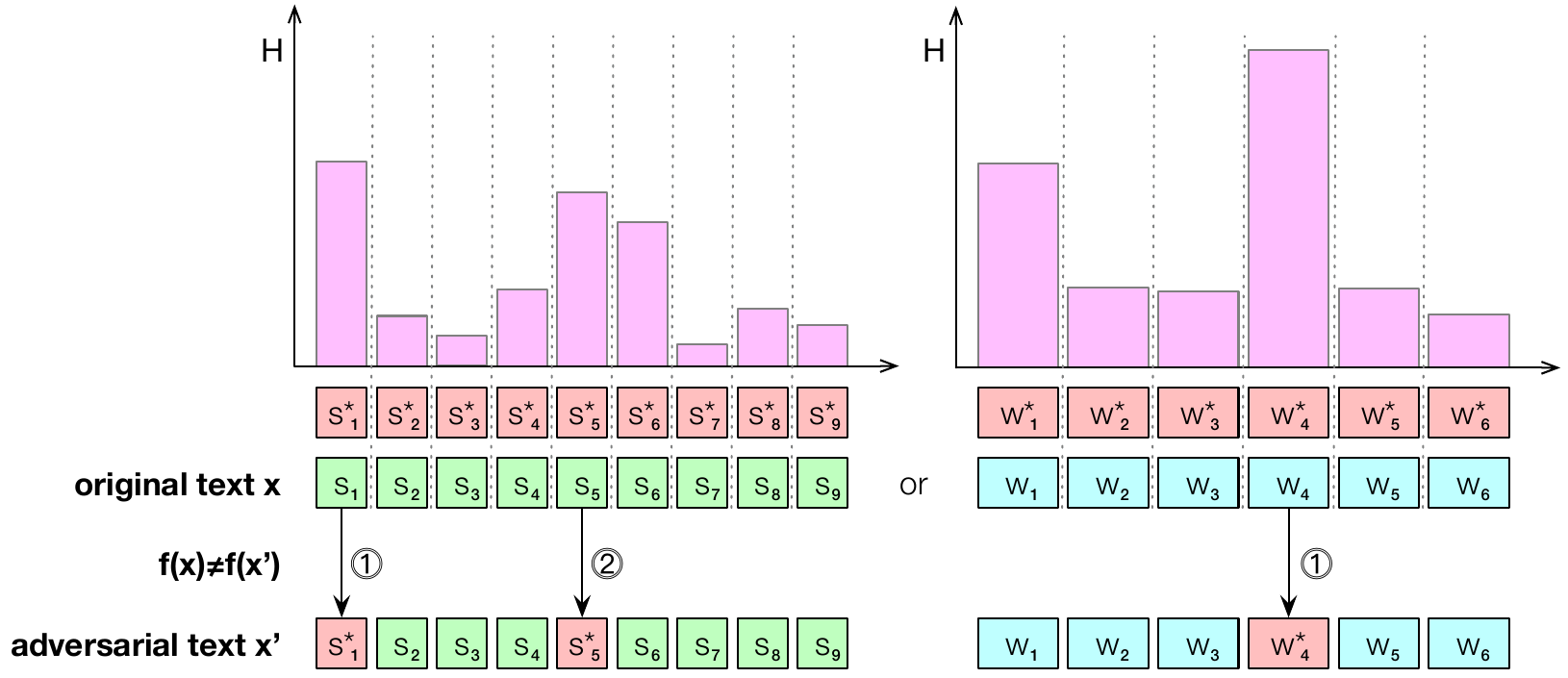}
	\caption{Process of determining substitution order.}
	\label{Fig.4}
\end{figure}

\subsection{Determine Substitution Order}

Word saliency \cite{li2017understandingneuralnetworksrepresentation} refers to the decrease in classification probability after setting a word in the original input text to unknown.
We use this metric to measure the saliency of a syllable or word.
Let
\begin{equation}
\hat{x}=s_{1}s_{2}\dots{unk.}\dots{s_{m}}~or~\hat{x}=w_{1}w_{2}\dots{unk.}\dots{w_{n}},
\end{equation}
\begin{equation}
S=P(y|x)-P(y|\hat{x}).
\end{equation}
Probability weighted word saliency \cite{ren-etal-2019-generating} is further proposed, which considers the saliency of both the substituted and the substitution word.
We use this metric to determine the substitution order.
The metric is defined as follows,
\begin{equation}
	H=Softmax(S)\cdot\Delta{P^{*}}.
\end{equation}
We sort all the scores $\{H_{1},H_{2},\dots\}$ corresponding to the original input text $x$ in descending order, then sequentially substitute the syllable $s_i$ with ${s_i}^*$ or the word $w_j$ with ${w_j}^*$ until the output label changes, which means the attack succeeds.
Otherwise, the attack fails.
Fig. \ref{Fig.4} depicts the process of determining substitution order.

\section{Experiments}

\subsection{PLMs}

\begin{itemize}
	\item
	\textit{Tibetan-BERT}\footnote{\url{https://github.com/UTibetNLP/Tibetan_BERT}} \cite{10.1145/3548608.3559255} is the first monolingual PLM targeting Tibetan which is based on BERT \cite{devlin-etal-2019-bert}.
	It achieves a good result in the specific downstream text classification task.
	\item
	\textit{CINO}\footnote{\url{https://github.com/iflytek/cino}} \cite{yang-etal-2022-cino} is the first multilingual PLM for Chinese minority languages including Tibetan which is based on XLM-RoBERTa \cite{conneau-etal-2020-unsupervised}.
	It achieves excellent results in multiple monolingual or multilingual downstream text classification tasks.
\end{itemize}

\subsection{Datasets}

\begin{itemize}
	\item \textit{TNCC-title}\footnote{\url{https://github.com/FudanNLP/Tibetan-Classification}} \cite{10.1007/978-3-319-69005-6_39} is a widely used Tibetan news title classification dataset.
	It contains 9,276 samples divided into 12 classes.
	\item \textit{TU\_SA}\footnote{\url{https://github.com/UTibetNLP/TU_SA}} \cite{MESS202302007} is a high-quality Tibetan sentiment analysis dataset proofread by humans.
	It contains 10,000 samples, of which negative or positive class each accounts for 50\%.
\end{itemize}

\subsection{Victim Language Models}

Eight victim language models are constructed via the paradigm of “PLMs + Fine-tuning”.
Each dataset is divided into a training set, a validation set, and a test set according to the ratio of 8:1:1.
On the training set and the validation set, we fine-tune the PLMs to construct the victim language models, and on the test set, we conduct the attack methods on the victim language models.
For a fair comparison, the fine-tuning hyper-parameters are taken from the previous work \cite{10.1145/3589335.3652503}.
We make all the Tibetan victim language models publicly available in our Hugging Face Collection\footnote{\url{https://huggingface.co/collections/UTibetNLP/tibetan-victim-language-models-669f614ecea872c7211c121c}}.

\subsection{Baselines}

\begin{itemize}
	\item \textit{TSAttacker} \cite{cao-etal-2023-pay-attention} is a syllable-level black-box Tibetan textual adversarial attack method.
	It utilizes cosine distance between static syllable embeddings to generate substitution syllables.
	\item \textit{TSTricker} \cite{10.1145/3589335.3652503} is a syllable- and word-level black-box Tibetan textual adversarial attack method.
	It utilizes two BERT-based masked language models with tokenizers of two different granularity to generate substitution syllables or words respectively.
	In TABLE \ref{Tab.2}, “-s” and “-w” represent syllable- and word-level attack respectively.
\end{itemize}

\begin{table*}[t]
	\centering
	\caption{Experimental Results\\\textbf{\underline{bold and underlined values}} represent \textbf{\underline{the best performance}}\\\textbf{bold values} represent \textbf{the second best performance}}
	\label{Tab.2}
	\begin{tabular}{cccccc|cccc}
		\toprule
		\multirow{2}{*}{Metric}&\multirow{2}{*}{Method}&\multicolumn{4}{c}{TNCC-title}&\multicolumn{4}{c}{TU\_SA} \\
		\cmidrule(r){3-6}\cmidrule(r){7-10}
		&& Tibetan-BERT & CINO-small & CINO-base & CINO-large & Tibetan-BERT & CINO-small & CINO-base & CINO-large \\
		\midrule
		\multirow{5}{*}{\textit{ADV} $\uparrow$} & TSAttacker & 0.3420 & 0.3592 & 0.3646 & 0.3430 & 0.1570 & 0.2260 & 0.2240 & 0.2660 \\
		& TSTricker-s & \textbf{\underline{0.5124}} & 0.5685 & 0.5414 & \textbf{\underline{0.5426}} & \textbf{\underline{0.3080}} & \textbf{\underline{0.4300}} & \textbf{\underline{0.4730}} & \textbf{0.5060} \\
		& TSTricker-w & \textbf{\underline{0.5124}} & 0.5588 & 0.5566 & 0.5286 &  \textbf{0.2870} & \textbf{0.4050} & 0.4200 & \textbf{\underline{0.5100}} \\
		& TSCheater-s & 0.4714 & \textbf{\underline{0.5717}} & \textbf{0.5620} & 0.5329 & 0.2810 & 0.3790 & 0.4390 & 0.4280 \\
		& TSCheater-w & \textbf{0.5027} & \textbf{0.5696} & \textbf{\underline{0.5696}} & \textbf{0.5405} & 0.2810 & 0.3770 & \textbf{0.4540} & 0.4260 \\
		\midrule
		\multirow{5}{*}{\textit{LD} $\downarrow$} & TSAttacker & 5.2000 & 5.6210 & 5.0638 & 5.3386 & 7.7298 & 7.4533 & \textbf{8.0769} & 7.3369 \\
		& TSTricker-s & 4.0671 & 5.8856 & 5.3402 & 5.6865 & \textbf{5.4887} & \textbf{6.2495} & 9.2057 & \textbf{6.8813} \\
		& TSTricker-w & 10.2492 & 13.0297 & 12.9511 & 12.3374 & 16.9542 & 14.2365 & 16.5066 & 16.7699 \\
		& \textbf{\underline{TSCheater-s}} & \textbf{\underline{1.6941}} & \textbf{\underline{2.5775}} & \textbf{\underline{2.5713}} & \textbf{\underline{2.8846}} & \textbf{\underline{2.1219}} & \textbf{\underline{3.5827}} & \textbf{\underline{4.7923}} & \textbf{\underline{3.5393}} \\
		& TSCheater-w & \textbf{3.1771} & \textbf{3.9363} & \textbf{4.0066} & \textbf{4.0531} & 7.4147 & 9.0522 & 10.0106 & 8.9047 \\
		\midrule
		\multirow{5}{*}{\textit{CS} $\uparrow$} & TSAttacker & \textbf{\underline{0.9653}} & \textbf{0.9644} & \textbf{0.9678} & \textbf{0.9666} & \textbf{\underline{0.9844}} & \textbf{0.9862} & \textbf{0.9841} & \textbf{0.9845} \\
		& TSTricker-s & \textbf{0.9602} & 0.9543 & 0.9603 & 0.9578 & 0.9750 & 0.9793 & 0.9739 & 0.9778 \\
		& TSTricker-w & 0.8865 & 0.8870 & 0.8895 & 0.8925 & 0.9315 & 0.9384 & 0.9316 & 0.9371 \\
		& TSCheater-s & 0.9547 & \textbf{\underline{0.9734}} & \textbf{\underline{0.9737}} & \textbf{\underline{0.9708}} & \textbf{0.9785} & \textbf{\underline{0.9903}} & \textbf{\underline{0.9865}} & \textbf{\underline{0.9908}} \\
		& TSCheater-w & 0.9447 & 0.9433 & 0.9433 & 0.9547 & 0.9417 & 0.9507 & 0.9501 & 0.9526 \\
		\midrule
		\multirow{5}{*}{\textit{VS} $\uparrow$} & TSAttacker & 0.5474 & 0.5633 & 0.5642 & 0.5783 & 0.5627 & 0.5848 & \textbf{0.5856} & 0.5943 \\
		& TSTricker-s & \textbf{0.5669} & 0.5720 & 0.5684 & 0.5748 & \textbf{0.5765} & \textbf{0.6203} & 0.5769 & \textbf{0.6583} \\
		& TSTricker-w & 0.3990 & 0.4280 & 0.4246 & 0.4401 & 0.3943 & 0.4032 & 0.3941 & 0.4097 \\
		& \textbf{\underline{TSCheater-s}} & \textbf{\underline{0.7145}} & \textbf{\underline{0.8152}} & \textbf{\underline{0.7967}} & \textbf{\underline{0.8113}} & \textbf{\underline{0.7262}} & \textbf{\underline{0.8177}} & \textbf{\underline{0.7758}} & \textbf{\underline{0.8370}} \\
		& TSCheater-w & 0.5540 & \textbf{0.6234} & \textbf{0.6110} & \textbf{0.6298} & 0.4679 & 0.4880 & 0.4892 & 0.4792 \\
		\bottomrule
	\end{tabular}
\end{table*}

\subsection{Evaluation Metrics}

\begin{itemize}
	\item \textit{Accuracy Drop Value} (\textit{ADV}) is the decrease of model accuracy post- compared to pre-attack, with which we evaluate the attack effectiveness. The larger the \textit{ADV}, the more effective the attack.
	\item \textit{Levenshtein Distance} (\textit{LD}) between the original text and the adversarial text is the minimum number of single-letter edits (insertions, deletions, or substitutions) required to change one into the other.
	We use this metric to evaluate the perturbation magnitude. The smaller the \textit{LD}, the more slight the perturbation.
	\item \textit{Cosine Similarity} (\textit{CS}) is the cosine of the angle between two vectors.
	Here, the calculation is based on the word embeddings of \textit{Tibetan-BERT}, and we use this metric to evaluate the semantic similarity between the original text and the adversarial text.
	The larger the \textit{CS}, the more semantically similar they are.
	\item \textit{CCOEFF\_NORMED} is used to evaluate the visual similarity between the original text and the adversarial text.
	Here, the calculation is the same as when constructing \textit{TSVSDB}.
	The larger the \textit{CCOEFF\_NORMED}, the more visually similar they are.
	In Table \ref{Tab.2}, it is abbreviated as \textit{VS}.
	\item According to the construction of AdvGLUE \cite{NEURIPS_DATASETS_AND_BENCHMARKS2021_335f5352} and AdvGLUE++ \cite{NEURIPS2023_63cb9921}, after human annotation, most existing adversarial attack algorithms are prone to generating adversarial examples that are difficult for humans to accept.
	After filtering out adversarial texts with $LD/len(orig\_text)$ less than 0.1, we also conduct human annotation.
	The rating of human acceptance for adversarial texts ranges from 1 to 5, and adversarial texts rated 4 or 5 are acceptable to humans.
	We utilize human-acceptable adversarial texts on Tibetan-BERT to construct the first Tibetan adversarial robustness evaluation benchmark called \textit{AdvTS}.
	TABLE \ref{Tab.3} presents the composition information of \textit{AdvTS}.
	From the table, we can see that \textit{TSCheater} generates the most human-acceptable adversarial texts.
	The two values in parentheses represent the number of human-acceptable adversarial texts generated by syllable- and word-level attack, respectively.
	\textit{AdvTS} is used to evaluate the adversarial robustness of models after \textit{Tibetan-BERT}.
	TABLE \ref{Tab.4} lists the evaluation results of \textit{CINO} on \textit{AdvTS}.
	We will continue to expand and maintain \textit{AdvTS} in the future.
\end{itemize}

\begin{table}[t]
	\centering
	\caption{Composition Information of AdvTS\\\textbf{\underline{bold and underlined values}} represent \textbf{\underline{the largest part}}}
	\label{Tab.3}
	\begin{tabular}{c|ccc|c}
		\toprule
		\diagbox{Dataset}{Method} & TSAttacker & TSTricker & \textbf{\underline{TSCheater}} & Total \\
		\midrule
		Adv-TNCC-title & 89 & 30 (19+11) & \textbf{\underline{226 (90+136)}} & 345 \\
		\midrule
		Adv-TU\_SA & 78 & 19 (10+9) & \textbf{\underline{151 (72+79)}} & 248 \\
		\bottomrule
	\end{tabular}
\end{table}

\begin{table}[t]
	\centering
	\caption{Adversarial Robustness of CINO on AdvTS}
	\label{Tab.4}
	\begin{tabular}{c|cc|c}
		\toprule
		\diagbox{Model}{Dataset} & Adv-TNCC-title & Adv-TU\_SA & Average \\
		\midrule
		CINO-small & 0.4928 & 0.6290 & 0.5609 \\
		\midrule
		CINO-base & 0.4812 & 0.6331 & 0.5572 \\
		\midrule
		CINO-large & 0.5362 & 0.6089 & 0.5726 \\
		\bottomrule
	\end{tabular}
\end{table}

\subsection{Experimental Results}

We conduct a comprehensive evaluation of the existing methods from five dimensions.
TABLE \ref{Tab.2} shows the performance in attack effectiveness, perturbation magnitude, semantic similarity, and visual similarity.
TABLE \ref{Tab.3} shows the performance in human acceptance.
\textit{TSCheater} utilizes the visual similarity of Tibetan syllables to generate adversarial texts, therefore the adversarial texts generated by \textit{TSCheater-s} have extremely high visual similarity with the original texts.
Since visually similar syllables are semantically similar to a certain extent, \textit{TSCheater-s} also performs very well in semantic similarity.
Visually similar syllables only differ slightly in the individual letters, and the perturbation of syllables is no longer a whole substitution but a partial modification, so \textit{TSCheater-s} achieves an extremely small perturbation magnitude.
We attribute the reason why \textit{TSCheater-w} does not perform as well as \textit{TSCheater-s} in these three dimensions to the larger perturbation granularity and the effect of the segmentation tool.
Textual adversarial attacks are more effective does not mean that the generated adversarial texts are more human-acceptable.
\textit{TSCheater} performs well in attack effectiveness and best in human acceptance.
Tibetan script is an abugida, and \textit{TSCheater} fully considers the encoding and textual features of this writing system, so our proposed method can also be transferred to other abugidas like Devanagari script.

\section{Conclusion}

In this study, we propose and open-source a fully covered Tibetan syllable visual similarity database called \textit{TSVSDB}, a simple and sweet Tibetan adversarial text generation method called \textit{TSCheater}, and the first Tibetan adversarial robustness evaluation benchmark called \textit{AdvTS}.
We believe \textit{TSCheater} has transferability in abugidas and we will continue to expand and maintain \textit{AdvTS}.
Finally, the adversarial robustness of language models covering low-resource cross-border languages plays a crucial role in AI and national security, so we call for more attention to it.

\section*{Acknowledgments}

This work is supported by the National Social Science Foundation of China (22\&ZD035), the Key Project of Xizang Natural Science Foundation (XZ202401ZR0040), the National Natural Science Foundation of China (61972436), and the MUC (Minzu University of China) Foundation (GRSCP202316, 2023QNYL22, 2024GJYY43).

\bibliographystyle{IEEEtran}
\bibliography{icassp2025}

\end{document}